\pdfoutput=1

\documentclass[11pt]{article}

\usepackage[]{emnlp2021}
\usepackage{times}
\usepackage{latexsym}
\usepackage[T1]{fontenc}
\usepackage[utf8]{inputenc}
\usepackage{microtype}

\usepackage{amsmath}
\usepackage{bm}
\usepackage{xspace}
\usepackage{graphicx}
\DeclareMathOperator*{\argmax}{argmax}
\newcommand{\g}{guiding text\xspace}
\newcommand{\G}{Guiding Text\xspace}
\newcommand{\gs}{guiding texts\xspace}
\newcommand{\Gs}{Guiding Texts\xspace}
\newcommand{\frcnnobj}{FRCNN\xspace}
\newcommand{\gcplabel}{GCP\xspace}
\newcommand{\best}[1]{\textbf{#1}}
\newcommand{\shortI}{$H_I$\xspace}
\newcommand{\shortC}{$H_C$\xspace}
\newcommand{\shortF}{$H_F$\xspace}
\newcommand{\IG}{$\mbox{G}$\xspace}
\newcommand{\IGGR}{$\mbox{G}$\xspace}
\newcommand{\IR}{$\mbox{R}$\xspace}
\newcommand{\IRFRCNN}{$\mbox{R}_\textsubscript{FRCNN}$\xspace}
\newcommand{\IRGR}{$\mbox{R}_\textsubscript{GR}$\xspace}
\newcommand{\GT}{$\mbox{T}$\xspace}
\newcommand{\pair}[2]{$\langle\mbox{#1}, \mbox{#2}\rangle$\xspace}
\newcommand{\tuple}{$\langle\mbox{image}, \mbox{\g}, \mbox{caption}\rangle$\xspace}
\newcommand{\data}[2]{$\mbox{#1}_\textsubscript{#2}$}
\newcommand{\smaller}[1]{\scriptsize{#1}\xspace}

\newcommand{\cider}{{\footnotesize{CIDEr}}\xspace}
\newcommand{\tablesize}{\footnotesize}

\title{Understanding Guided Image Captioning Performance across Domains}

\author{ \\
  Edwin G. Ng\thanks{\; Work done as a part of the Google AI Residency.} \quad Bo Pang \quad Piyush Sharma \quad Radu Soricut \\
  Google Research \\
  \texttt{eg.ng@alum.utoronto.ca, \{bopang,piyushsharma,rsoricut\}@google.com} \\}

\begin{document}
\maketitle
\begin{abstract}
Image captioning models generally lack the capability to take into account user interest, and usually default to global descriptions that try to balance readability, informativeness, and information overload.
We present a Transformer-based model with the ability to produce captions focused on specific objects, concepts or actions in an image by providing them as \g to the model.
Further, we evaluate the quality of these guided captions when trained on Conceptual Captions which contain 3.3M image-level captions compared to Visual Genome which contain 3.6M object-level captions.
Counter-intuitively, we find that guided captions produced by the model trained on Conceptual Captions generalize better on out-of-domain data.
Our human-evaluation results indicate that attempting in-the-wild guided image captioning requires access to large, unrestricted-domain training datasets, and that increased style diversity (even without increasing the number of unique tokens) is a key factor for improved performance.
\end{abstract}

\section{Introduction}
Describing the content of an image using natural language is generically referred to as image captioning, but there are a variety of ways in which this can be achieved: by focusing on the most salient aspects of an image, as in MSCOCO~\cite{coco} or Conceptual Captions~\cite{sharma2018conceptual}; on most of the groundable concepts in an image, as in Image Paragraphs~\cite{krause2016paragraphs} or Localized Narratives~\cite{PontTuset_eccv2020}; or on a predefined set of objects, as in dense captioning~\cite{johnson2016densecap}.
These various approaches acknowledge that a typical real-world image may contain a varying number of objects/concepts/actions that may be of interest to the caption consumer, and therefore the optimal description depends on the degree to which the caption covers what the user is interested in at any given moment.

\begin{figure}[t]
    \centering
    \includegraphics[width=1.0\columnwidth]{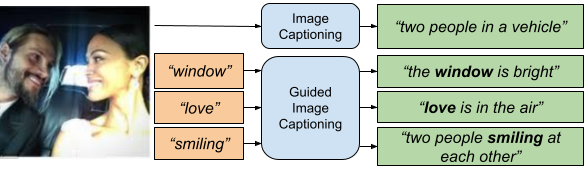} \\
\caption{
    An illustration comparing the difference between an image captioning model and a guided image captioning model. The guided image captioning model generates captions that focus on a specific object, concept or action of the image, provided to the model as free-form \g.
}
\label{fig:intro}
\end{figure}

However, the vast majority of captioning solutions lack the ability to take into account user interest, and usually default to global descriptions.
Such a description might say, for example \textit{``people attending a birthday party''}, which
may not satisfy a user interested in a description of \textit{``the cake''}.
In this paper, we capture user interest via the {\em \g}, a free-form text input that is assumed to be related to some concept(s) in the image; and
we consider the {\em Guided Image Captioning} task, 
where the \g is provided to the model as an additional input to control the concepts that an image caption should focus on\footnote{A key difference to the Visual Question Answering (VQA) task \citep{anderson18bottomup, zhou20unified} is that Guided Image Captioning is framed as a generation task to produce an open-ended description, while VQA is usually framed as a classification task to produce a specific closed-ended answer.
In its most frequent approaches, VQA uses a long-text input form (the question) and a short-text output form (the answer). In contrast, Guided Image Captioning uses a short-text input form (the guiding text) and a long-text output form (the caption).}.
This could, for instance, enable accessibility tools for visually-impaired users, who can select a guiding text produced from an upstream object/label detector to receive a guided description of their surroundings.
Note that \gs are not limited to a set of boxable objects, but include concepts (e.g. \textit{``vacation''}, \textit{``love''}) and actions (e.g. \textit{``swimming''}).

A key question we ask in this work is what kind of training data can lead to models that work better in a real-world setting.
Ideally, the training data should contain multiple captions for a given image, each associated with a different \g. 
The Visual Genome dataset~\cite{krishna-etal-2017-dataset} provides exactly that, with a total of 3.6M object-level captions created through human annotation.
In contrast, the Conceptual Captions~\cite{sharma2018conceptual} contains only image-level captions created  through an automatic pipeline obtaining images and captions from the internet.
We perform human evaluations that measure caption informativeness, correctness and fluency to test our approach in a real-world setting. 
Interestingly, while the Conceptual Captions dataset contains a similar number of captions (3.3M), and has a lower number of unique tokens than Visual Genome, models trained on this dataset generalize better on out-of-domain data in human evaluations.

The key contributions of this paper are summarized as follows:
\begin{itemize}
    \item Given the popularity of Transformer~\cite{vaswani2017attention} models for generative tasks in NLP, we use a multimodal Transformer model as a vehicle to study characteristics of the guided image captioning task.  In particular, we investigate the underlying characteristics of image captioning datasets required to produce higher quality guided captions.  Our results suggest that the key to solving in-the-wild guided image captioning may not be laborious human annotations, Visual Genome--style, but rather access to noisy, unrestricted-domain training datasets with high style diversity. 
    \item We open-source the set of test images and \g pairs used for our human evaluation experiments in order to encourage future work in this direction and facilitate direct comparisons with our results.
\end{itemize}

\section{Related Work}
The task of image captioning has received considerable interest over the past decade \citep{chen2015mind,kiros2014multimodal,zhu2018captioning,donahue2015long, mao15mrnn, karpathy2014deep, vinyals2015show}. 
Template-based methods \citep{kulkarni2013babytalk,elliott2015describing,devlin2015language} provide image grounding, but lack the ability to produce diverse captions, while early encoder-decoder methods \citep{donahue2015long,vinyals2014show,karpathy2014deep, zheng2019intention,wu2015image, sharma2018conceptual} produce diverse captions, but lack image grounding.
Contemporary encoder-decoder methods \citep{lu18neural, cornia2019show, anderson2017guided,you2016image,chen2020say,mun2017text,xu15show,fang2015captions}, including the method proposed in this work, can generate diverse captions with image grounding.

Most similar to our work, ~\cite{zheng2019intention} propose the use of a guiding object and produces a caption by using a forward and backward LSTM to separately generate the caption text before and after the guiding object.
In contrast, our approach involves a multi-modal Transformer model that uses layers of self-attention and cross-attention to better ground the target caption using early-fused representations for the image and the \g, which in our work is not limited to boxable objects.

\section{Method}
\label{sec:method}
Given an image $\bm{I}$ and \g $\bm{T}=\{t_1,...,t_{L_T}\}$ with $L_T$ tokens, the goal of Guided Image Captioning is to generate an image description $\bm{y}=\{y_1,...,y_{L_y}\}$ with $L_y$ tokens, such that $\bm{y}$ focuses on the concepts provided in $\bm{T}$. Note that we do not necessitate our model to include the tokens of $\bm{T}$ inside $\bm{y}$, yet we find that our trained model produces $\bm{T}$ inside $\bm{y}$ a majority of the time. Interestingly, in some cases where $\bm{T}$ is absent from $\bm{y}$, we find that $\bm{T}$ is paraphrased in $\bm{y}$. See Figure \ref{fig:para_incorrect_samples} for examples.
Both $t_i$ and $y_i$ are tokens in the shared model vocabulary.

We employ a Transformer \citep{vaswani2017attention} based sequence to sequence model, with parameters $\bm{\theta}$ where $\bm{I}$ and $\bm{T}$ are inputs to the encoder and $\bm{y}$ is the decoder output.
We use a dataset with $(\bm{I}, \bm{T}, \bm{y})$ tuples to search for the optimal model parameters $\bm{\theta^*}$ by maximizing the likelihood of the correct caption given below.

\begin{equation}
    \bm{\theta^*}=\argmax_{\bm{\theta}} \sum_{(\bm{I},\bm{T},\bm{y})}\log{p(\bm{y}|\bm{I},\bm{T};\bm{\theta})}
\end{equation}

In our model, the encoder input sequence comprises of image features followed by sub-token embeddings for the \g.
We use a joint encoder for both image and \g to transform the image features in the context of the \g concepts and vice versa, such that the decoder input is conditioned on the joint text-image representation.

We represent the input image as a sequence of global and regional features, \IG and \IR, respectively.
Following \citet{changpinyo2019decoupled}, we use Graph-Regularized Image Semantic Embedding (GraphRISE) as the global image features \IGGR.
These 64-dimensional features, trained to discriminate O(40M) ultra-fine-grained semantic labels, have been shown to outperform other state of the art image embeddings \cite{juan2019graphrise}.

For regional features, we first compute bounding boxes using a Region Proposal Network (RPN).  Following \citet{anderson18bottomup}, we reimplement the Faster R-CNN (FRCNN) model trained to predict selected labels
(1,600 objects and 400 attributes) in Visual Genome \cite{krishnavisualgenome}.
This model returns $K$ bounding box regions, each with a 2048-dimensional feature vector.
\citet{changpinyo2019decoupled} observed that decoupling box proposal and featurization could improve downstream tasks.
Thus, we obtain two sets of regional features -- \IRGR~and \IRFRCNN~where the bounding box feature extractor are GraphRISE and FRCNN respectively.
We experimented with using either or both of these features  for the top 16 regions.

\subsection{Encoder}
\begin{figure}[t]
    \centering
    \includegraphics[width=0.75\columnwidth]{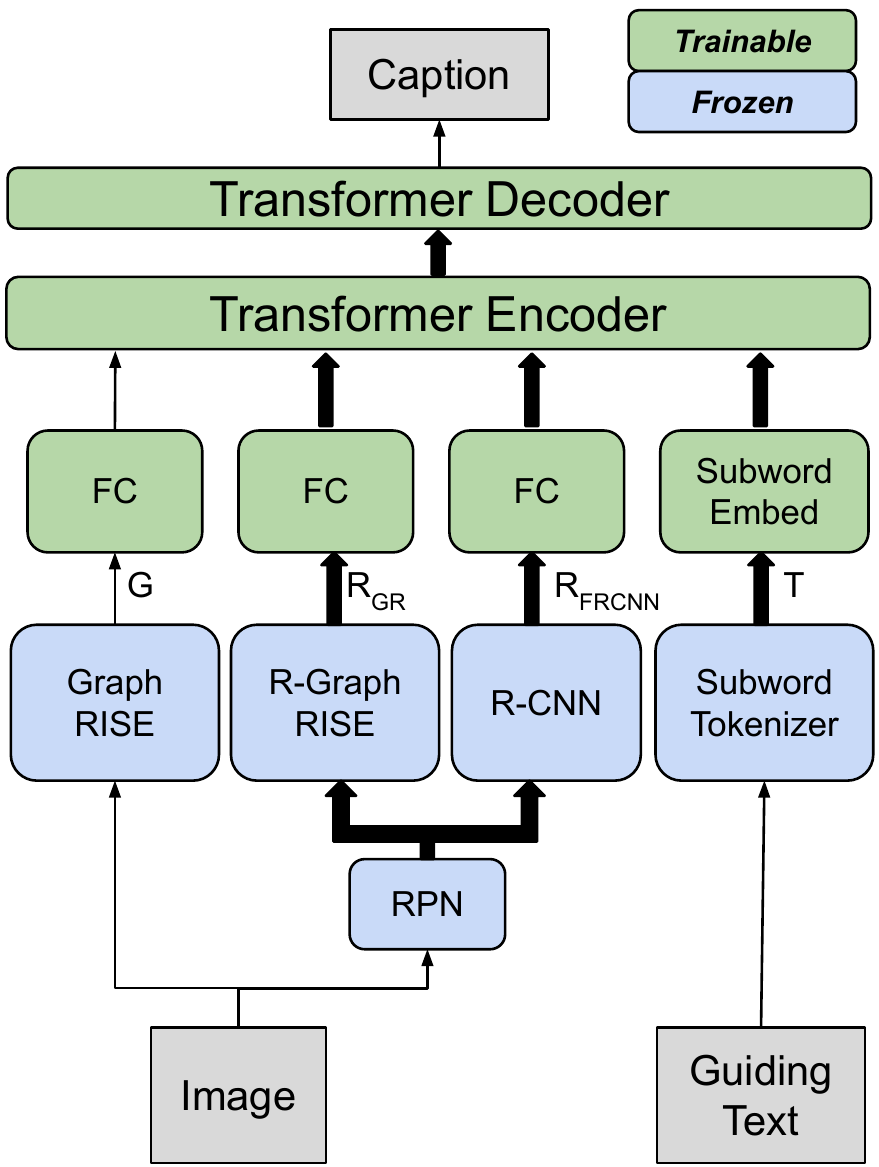}
\caption{
    Our model architecture. 
    Thick arrows indicate a sequence, such as bounding boxes for image regions, or list of tokens in a text.}
\label{fig:model}
\end{figure}

All models in the image feature extraction pipeline, are pre-trained and are not fine-tuned during our model training process. 
To transform them into the encoder input space, we use a trainable fully-connected network for each type of feature, denoted as $FC$ in Figure~\ref{fig:model}. 
The \g is provided to the encoder as a sequence of sub-tokens, \GT, which are embedded in the input space using trainable embeddings.

To summarize, the input to the Transformer encoder is a sequence of features in the following order.

\begin{description}
\itemsep0em
\item[\IGGR:] Global image features by Graph-RISE, 64D vector.
\item[\IRGR:] Sequence of 16 regional image features, 64D each, obtained by running GraphRISE on each image region.
\item[\IRFRCNN:] Sequence of 16 regional image features, 2048D each, obtained from FRCNN for each image region.
\item[\GT:] Sequence of sub-tokens corresponding to the \g.
\end{description}

\subsection{Decoder}
The Transformer decoder generates the output caption sequence $\bm{y}$ conditioned on the encoder outputs. 
The decoder shares the same vocabulary and embeddings used for the \g by the encoder.

\section{Data}
\label{sec:data}

We train our models on two datasets with different characteristics, we pick models according to automatic metrics computed over the validation (dev) split in each dataset.
Human evaluations were conducted on a separate test set which is out-of-domain for both  training datasets. The sub-token vocabulary size is set to 4000 for both datasets, but a comparison of the number of unique tokens is reported in Table \ref{tab:data}.

\subsection{Training and Validation (Dev) data}
\paragraph{Visual Genome (VG).} Visual Genome \cite{krishnavisualgenome} contains dense annotations for each image, including a list of human-annotated objects and {\em region captions} (descriptions of image regions localized by a bounding box).
We take all region captions and their first associated object to form \pair{\g}{caption} pairs for each image.
On average, each image has 34 such pairs, providing 3.6M \tuple tuples for 107,180 images. We randomly sample 96,450 and 10,730 images for the training and validation (dev) sets respectively.

\paragraph{Conceptual Captions (CC).}
Conceptual Captions \cite{sharma2018conceptual} contains 3.3 million training, 15K validation image/caption pairs collected from the internet.
Captions in this dataset are obtained from the alt-text of the images, and are diverse in their writing styles, however no paired object annotations are available for these captions.
Instead, we use the Google Cloud Natural Language API to extract the text span considered to be the most salient\footnote{Salience measures the importance of an entity in a caption (e.g. 0.65 for “cottage” and 0.35 for “seaside town” with “thatched cottage in the seaside town” as the caption).} entity in a caption, and treat it as the \g for the corresponding caption, yielding one \pair{\g}{caption} pair per image.

\begin{table}[t]
\tablesize
\begin{center}
\begin{tabular}{l|c|c} \hline
             & CC   & VG \\ \hline
\# of images & 3.3M & 107K \\
\# of \tuple & 3.3M & 3.6M \\
\# of unique \gs & 43.2K & 75.4K \\
\# of unique tokens & 42.6K & 69.5K \\ \hline
\end{tabular}
\caption {Statistics on the two datasets: Conceptual Captions (CC), Visual Genome (VG).}
\label{tab:data}
\end{center}
\end{table}

\begin{table}[t]
\tablesize
\centering
\begin{tabular}{c|rrr} \hline
     Length of guiding text & 1 & 2 & $\ge$ 3 \\ \hline
     CC &  80.5\% & 16.3\% & 3.4\% \\
     VG & 91.1\% & 8.3\% & 0.6\% \\ \hline
\end{tabular}
\caption{Length (in number of tokens) distribution of guiding texts in the two datasets.}
\label{tab:guiding_text_len}
\end{table}

\begin{table}[t]
\tablesize
\begin{center}
\begin{tabular}{l|rrr|rrr} \hline
Test T     &   \multicolumn{3}{c|}{unique \gs}     &   \multicolumn{3}{c}{unique tokens} \\
source & cnt & $\in$ \smaller{CC} & $\in$ \smaller{VG} & cnt & $\in$ \smaller{CC} & $\in$ \smaller{VG} \\ \hline \hline
\smaller{\gcplabel} & 889 & 67\% & 62\% & 929 & 90\% & 88\% \\ 
\smaller{\frcnnobj} & 421 & 98\% & 100\% & 421 & 99\% & 100\% \\ \hline 
\end{tabular}
\caption {Percentage of unique (tokens in) test guiding texts that have been seen at training time for Conceptual Captions (CC) and Visual Genome (VG).}
\label{tab:overlap}
\end{center}
\end{table}

\paragraph{Comparisons of the two datasets.}
As shown in Table \ref{tab:guiding_text_len}, the \gs are dominated by single-word expressions with a minority of multi-word expressions (19.5\% for CC; 8.9\% for VG). Both datasets have a similar number of \tuple tuples.
While VG has significantly fewer images, it has a much higher number of unique \gs and its number of unique tokens in \gs and captions combined is larger than that of CC (Table \ref{tab:data}).

Which type of dataset is more suitable to train a guided image captioning model?
The VG dataset contains human-quality \gs and corresponding captions, even for small regions in the image and also has a higher number of unique tokens.
In contrast, the CC dataset contains noisy \gs automatically extracted from captions, which potentially focus only on the most salient concept in the image.
However, it has more image diversity, and its captions reflect more diverse styles as a result of a much larger set of authors. 
Although the model trained with CC is exposed to only one \pair{\g}{caption} pair per image compared to VG which has on average 34 \pair{\g}{caption} pairs per image, it may still learn to correctly generate descriptions for different concepts in an image, because it has been exposed to different ways of describing such images.

\begin{table*}[t]
\tablesize
\begin{center}
\begin{tabular}{l|l|cccc|rc}
\hline
    &      &   \multicolumn{4}{c|}{Automatic Metrics on \textbf{Dev Set}}     &   \multicolumn{2}{c}{Human Score on \textbf{T2}} \\
Train/Dev  &  Model                               & \smaller{CIDEr} & \smaller{SPICE} & \smaller{ROUGE-L} & \smaller{METEOR}  &   \smaller{GCP}   & \smaller{FRCNN} \\
\hline \hline
CC         & Set output to \GT                     & 0.580        & 0.301        & 0.204      & 0.087          &   -     &   -   \\
           & \GT                                   & 1.079        & 0.250        & 0.301      & 0.147          &   -     &   -   \\
           & \IGGR                                 & 0.808        & 0.171        & 0.242      & 0.117          &   -     &   -   \\
           & \GT + \IGGR                           & 1.683        & 0.337        & 0.383      & 0.204          &   -     &   -   \\
           & \GT + \IGGR + \IRFRCNN                & 1.685        & 0.337        & 0.383      & 0.205          &   -     &   -   \\
           & \GT + \IGGR + \IRGR                   & \textbf{1.718} & \textbf{0.341} & \textbf{0.387} & \textbf{0.208}          &   \best{0.73}     &   0.75   \\
           & \GT + \IGGR + \IRGR + \IRFRCNN        & 1.695        & 0.339        & 0.387      & 0.206          &   0.69     &   0.75   \\
\hline \hline
VG         & Set output to \GT                    & 1.108        & 0.372        & 0.277      & 0.119          &   -     &   -   \\
           & \GT                                  & 1.589        & 0.344        & 0.336      & 0.186          &   -     &   -   \\
           & \IGGR                                & 0.406        & 0.109        & 0.183      & 0.079          &   -     &   -   \\
           & \GT + \IGGR                          & 1.606        & 0.355        & 0.340      & 0.190          &   -     &   -   \\
           & \GT + \IGGR + \IRFRCNN               & 1.758        & 0.376        & 0.357      & 0.204          &   -     &   -   \\
           & \GT + \IGGR + \IRGR                  & 1.744        & 0.373        & 0.354      & 0.202          &   0.62     &   0.79   \\
           & \GT + \IGGR + \IRGR + \IRFRCNN       & \textbf{1.775} & \textbf{0.377} & \textbf{0.362} & \textbf{0.206}          &   \best{0.66}     &   \best{0.84}   \\
\hline
\end{tabular}
\caption {Automatic metrics and Human evaluation results for model variations trained on Conceptual Captions (CC) and Visual Genome (VG).
    Each model corresponds to a different set of inputs to the Transformer encoder.
    The first row for each dataset is a ``no-op model'' that simply copies the \g to output to compute baseline metrics.
    Human evaluations are conducted with two sets of \gs obtained for the test images -- from Google Cloud API (GCP) and an FRCNN model.
    In all but one case, the model with higher CIDEr score (note that CIDEr score is only comparable within the same dataset) is also the one receiving better average human ratings.}
\label{tab:auto-and-human-evals}
\end{center}
\end{table*}

\subsection{Test data}
\label{sec:data:test}
\paragraph{Images} For human evaluation, we use the T2 test set (i.e. 1000 images sampled from the Open Images Dataset \cite{oidv4_journal} for the Conceptual Captions challenge) and these images are out-of-domain for both training datasets. 

\paragraph{\Gs}
The T2 dataset does not provide any groundtruth annotations.
However, we want to evaluate models on {\em multiple} \gs for each image, in order to test their ability to generate different captions.
To this end, we compile a list of up to 6 \gs for each T2 test image as described below, which we release for reproducibility.
This is intended to simulate the scenario where a visually-impaired user is given a set of candidate \gs generated by an upstream model, and picks the one they are interested in learning more about. The \g chosen by the user is then provided as input to our model.
Specifically, we use the following two approaches, and consider up to 3 test \gs per image from each.

\begin{figure}[ht]
\centering
\includegraphics[width=\linewidth]{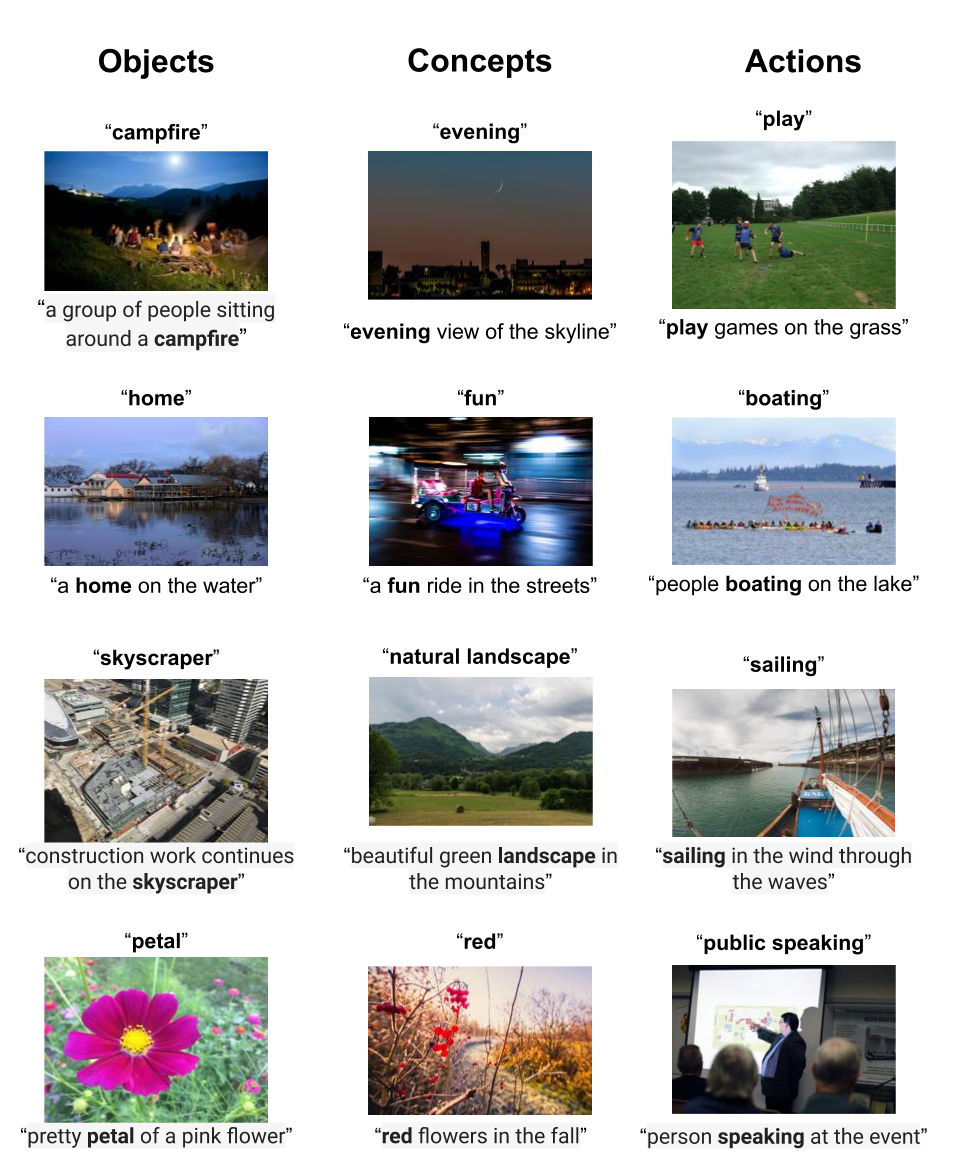}
\caption{Sample outputs for T2 set using the model trained on Conceptual Captions with GCP labels extracted as guiding text and evaluated on GCP labels. The first, second and third columns display samples with object-based, concept-based and action-based guiding text respectively.}
\label{fig:samples}
\end{figure}

\paragraph{GCP}
Three GCP labels are obtained using the image labelling API from Google Cloud by sorting labels by confidence scores and randomly sampling one from each tertile\footnote{We hope to evaluate our model performance over a wider variety of \gs from the upstream models, rather than limiting to the most confident ones.}.
The GCP labels generated for the T2 set can be an object (i.e. a box can be drawn around it to show its location in the image), a concept (a box cannot be drawn around it) or an action (i.e. a verb). Through manual inspection we find that 53.43\% are objects (e.g. “airplane”, “coin”, “desk”), 41.73\% are non-object concepts (e.g., “love”, “adventure”, “software engineering”) and 4.84\% are actions. 
Figure \ref{fig:samples} displays example GCP labels as well as sample outputs of each type of guiding text. 

\paragraph{FRCNN}
The other three guiding texts are FRCNN objects obtained from the faster R-CNN model described in Section \ref{sec:method} by also sorting the objects based on confidence and randomly sampling one from each tertile.
Since the model is trained on VG annotations, and its region proposal is used in our image representations, this simulates a much more in-domain type of \gs.
As expected, all of the FRCNN-based \gs have appeared in the training \gs of VG (Table \ref{tab:overlap}).

Overall, GCP-based \gs have a significantly lower overlap with our training data (for both CC and VG), as compared to FRCNN-based \gs.
We find that 67\% and 62\% of the GCP labels are seen at training time for CC and VG respectively, while 98\% and 100\% of the FRCNN objects are seen at training time for CC and VG respectively.
We consider our analysis with FRCNN objects as an in-domain type of \g and GCP labels as an out-of-domain type of \g.

This set of images, together with the \g used in our experiments, is made available\footnote{Download at \url{https://github.com/google-research-datasets/T2-Guiding}} in order to encourage future work in this direction and facilitate direct comparisons with our work.

\section{Experiments}
\label{sec:results}
\subsection{Model Implementation Details}
Our Transformer contains a stack of 6 layers each for both the encoder and decoder, with 8 attention heads.
All models were optimized with a learning rate of 0.128 using stochastic gradient descent (SGD) and a decay rate of 0.95. 
Decoding was performed using beam search with a beam width of 5.
The \GT + \IG + \IRGR + \IRFRCNN model has a total of 48.6M trainable parameters, while the \GT + \IG + \IRGR model has 47.3M trainable parameters.
We use a batch size of 4096 on a 32-core Google Cloud TPU.
All models are trained to optimize the \cider performance on the validation set, which typically takes approximately 2 million and 4 million steps for Visual Genome and Conceptual Captions models, respectively. 
With this setup, convergence for Visual Genome models typically takes 3 days, while Conceptual Captions takes 6 days.

We start from the initial hyperparameters from \cite{changpinyo2019decoupled}, and perform an additional hyperparameter search for the learning rate and decay rate for SGD.
The values used for the learning rate are \{0.0016, 0.008, 0.016, 0.048, 0.096, 0.128, 0.16\}, while for the decay rate are \{0.90, 0.95\}.
We choose the learning rate and decay rate with the highest validation CIDEr score (i.e. 0.128 learning rate and 0.95 decay rate), and use the same hyperparameters for all models.

\subsection{Automatic evaluation results}
We start by examining model performances via automatic evaluation metrics.
The CC-trained model is evaluated on the CC validation set, and the VG-trained model is evaluated on the VG validation set.
Note that the automatic evaluation metric results are only comparable within each dataset.

As shown in Table \ref{tab:auto-and-human-evals}, providing \gs as additional input to the model outperforms the model with image-only input (model \GT + \IG vs model \IG) for both datasets. 
This confirms the expectation that with either dataset, the model does learn to produce captions based on the \gs.
Model \GT + \IG also outperforms a trivial baseline method where the prediction reproduces the \g verbatim (``Set output to T''), 
and a stronger baseline, where only the \g is given as input to the model (Model \GT + \IG vs model \GT), confirming that the model does learn from the global image features as well.
Note that the VG-trained model achieves much higher \cider scores with the \GT baseline, indicating that the dataset contains more image-independent, formulaic expressions (e.g. ``white clouds in blue sky'').
Also, note that the a VG image has several (guiding text, image, caption) triples; with the guiding text removed, the \IG model is trained on multiple (image, caption) pairs, i.e. multiple targets for the same input, potentially causing the performance to be lower. 

The most important result here is that, for both datasets, the models that use both image features and \g achieve significantly higher scores than the \GT baseline,
indicating that these models are capable of learning additional, image-specific information about the \g.
Note that we observe a bigger improvement by adding image features to model \GT when training on the CC data (1.079 $\to$ 1.718 = +0.639 \cider), compared to the VG dataset (1.589 $\to$ 1.775 = +0.186 \cider).

Table~\ref{tab:auto-and-human-evals} also shows the effects of different visual features.
\IRFRCNN has a stronger positive effect on the Visual Genome dataset, most likely because the model that produces these features is trained on Visual Genome data.  The top two performing models for each dataset were sent for human evaluation, and we found that the human evaluation results correlated with the automatic metrics.

\begin{figure*}
    \centering
    \includegraphics[width=\linewidth]{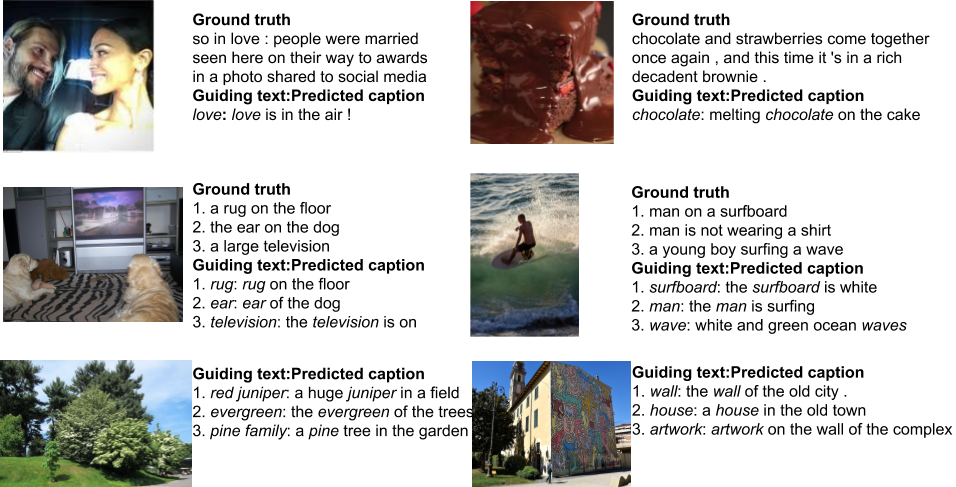}
\caption{Example qualitative results for CC images using model \GT + \IGGR + \IRGR trained on CC (first row), VG images using model \GT + \IGGR + \IRGR + \IRFRCNN trained on VG (second row), T2 images using model \GT + \IGGR + \IRGR trained on CC where guiding text is extracted using the Google Cloud Natural Language API (third row).}
\label{fig:qualitative}
\end{figure*}

See Figure \ref{fig:qualitative} for qualitative examples of models trained with different datasets. From Figure \ref{fig:qualitative}, we observe that although CC target captions are generally longer, our model naturally generates shorter captions even though we do not restrict the number of output tokens in the output caption.

\subsection{Human Evaluations}
\label{sec:human-eval}

For human evaluations over T2 images, each predicted caption is rated by three raters. Our rater pool consisted a total of 12 raters. Raters are shown an image, a guiding text, along with a generated image caption.  If the guiding text is judged to be present in the image\footnote{We find that automatically extracted T2 \gs using GCP and FRCNN were rated present in the image for 80.90\% and 79.51\% cases respectively.}, the rater is asked to rate the predicted caption across the following three dimensions; their choice will be converted to a score between 0 and 1 according to the following scheme:

\paragraph{Informativeness}: For someone who cannot see the image, does the caption provide additional information about the object?
\vspace{-.5em}
\begin{itemize}
\itemsep-0.5em
    \item No useful info provided [0.0]
    \item Some useful info provided [0.5]
    \item Key info provided [1.0]
\end{itemize}

\paragraph{Correctness}: Is the additional information correct?
\vspace{-.5em}
\begin{itemize}
\itemsep-.5em

    \item Incorrect  [0.0]
    \item Partially correct  [0.5]
    \item Correct   [1.0]
\end{itemize}

\paragraph{Fluency}: Does the caption read fluently?
\vspace{-.5em}
\begin{itemize}
\itemsep-.5em

    \item Not fluent. [0.0]
    \item Fluent [1.0]
\end{itemize}

\begin{table}[tb]
\tablesize
\begin{center}
\setlength\tabcolsep{3pt}
\begin{tabular}{c|ccc|c|cc|cc}
\hline
& \multicolumn{4}{c|}{Human score on \textbf{T2}} & \multicolumn{2}{c|}{Diversity} & \multicolumn{2}{c}{T $\in$ cap?}\\ 
$\mathcal{D}$ & \shortI & \shortC & \shortF & Avg & Div-1 & Div-2 & = \% & $\approx$ \% \\ \hline \hline
\multicolumn{9}{c}{Test $T$ = \gcplabel} \\ \hline
CC & 0.63 & 0.68 & 0.88 & 0.73 & 0.74 & 0.92 & 83.9 & 12.5\\
VG & 0.57 & 0.61 & 0.79 & 0.66 & 0.78 & 0.94 & 55.7 & 15.0 \\ 
MX & 0.70 & 0.68 & 0.88 & 0.75 & 0.78 & 0.94 & 86.7 & 11.5 \\ \hline \hline
\multicolumn{9}{c}{Test $T$ = \frcnnobj} \\ \hline
CC  & 0.65 & 0.67 & 0.94 & 0.75 & 0.76 & 0.94 & 98.3 & 0.2 \\
VG  & 0.75 & 0.80 & 0.96 & 0.84 & 0.74 & 0.92 & 99.8 & 0.2 \\
MX  & 0.74 & 0.72 & 0.95 & 0.80 & 0.78 & 0.95 & 99.5 & 0.1\\ \hline

\end{tabular}
\caption {Human evaluation results on T2 with the best models from each dataset ($\mathcal{D}$): Conceptual Captions (CC), Visual Genome (VG) and mixing both (MX). 
    Each instance was replicated three times and rated against informativeness (\shortI), correctness (\shortC) and fluency (\shortF); we report average score across these three scales ranging from 0 to 1. We report the n-gram diversity for uni-gram (Div-1) and bi-gram (Div-2). Last two columns report the percentage cases where the input \g T appears verbatim (=) or paraphrased ($\approx$) in the output caption.}
\label{tab:human-eval}
\end{center}
\end{table}

\begin{figure}[ht]
\centering
\includegraphics[width=\linewidth]{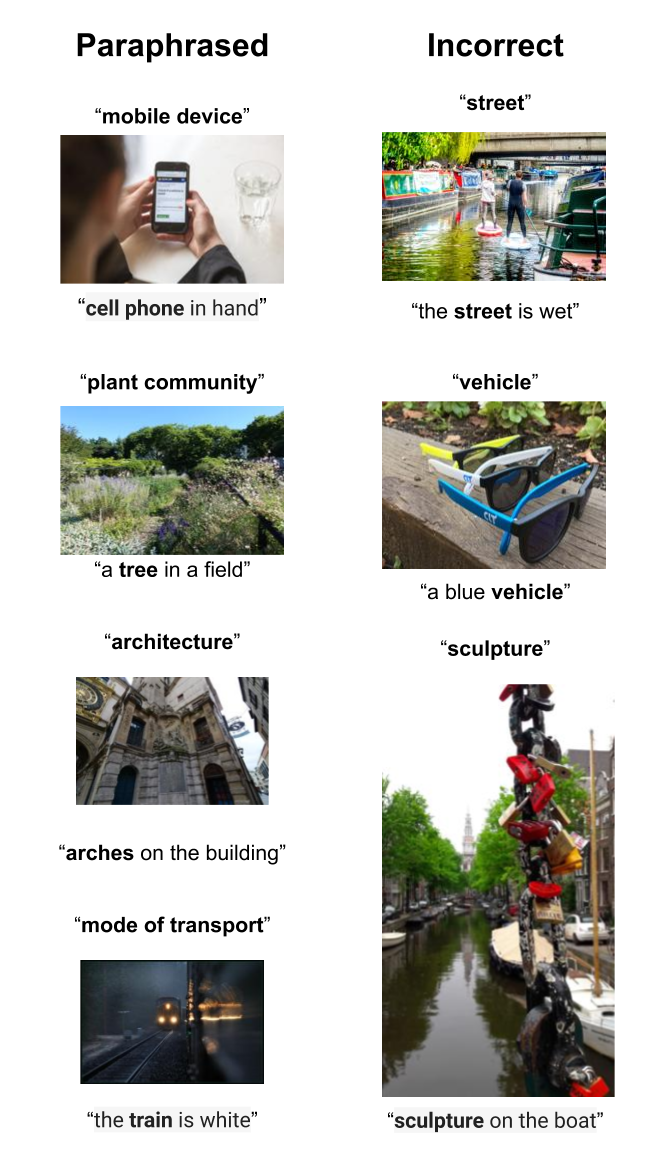}
\caption{Examples of captions that paraphrased the guiding text (first column) and examples of captions produced from guiding text that are absent from the image (second column). Sample outputs are obtained on the T2 set using the model trained on Visual Genome with its own annotations as guiding text and evaluated on GCP labels.}
\label{fig:para_incorrect_samples}
\end{figure}

Table~\ref{tab:human-eval} reports the average scores per dimension ranging from 0 to 1.
We observe that the VG-trained model performs significantly worse than the CC-trained model for out-of-domain GCP-based \gs.
One may hypothesize that the VG-trained model performs worse, because it contains fewer images than CC.
However, from Table~\ref{tab:human-eval} we also observe that the VG-trained model performs better than the CC-trained model for in-domain FRCNN-based \gs.
This suggests that the VG-trained model has a harder time adapting to the out-of-domain \gs, rather than the out-of-domain images.
But recall from Section \ref{sec:data:test}, the GCP-based \gs have similar levels of overlap with these two datasets, so why does the VG-trained model perform worse?
One potential reason could lie in the distribution of the \gs: entropy of \gs is significantly lower for VG with $H(T)$ = 9.7 bits, vs $H(T) $= 10.6 bits for CC, even though VG has a higher number of unique \gs and a higher number of unique tokens.
We posit that variety in the training data is important when adapting to out-of-domain \gs. 

\subsection{Caption Diversity}
Recall that we produce three captions for each image given three different \gs.
In Table \ref{tab:human-eval}, 
we report the n-gram diversity \cite{deshpande2019fast} for unigrams (Div-1) and bigrams (Div-2) by computing the number of distinct n-grams from the three predicted captions divided by the total number of n-grams from these three captions.  
Our Div-1 ranges 0.74 - 0.78, and Div-2 ranges from 0.92 to 0.95, showing that our model generates diverse captions for the same image when prompted by different guiding texts.

\subsection{\G paraphrasing in Caption}
Table \ref{tab:human-eval} also reports the percent of instances where the guiding text is present verbatim or paraphrased in the predicted caption, based on manual inspection.
Interestingly, we observe cases where paraphrasing the \g leads to a better caption.
For example: ``mobile device'' to ``cell phone in hand'', ``red juniper'' to ``a huge juniper in a field'', and ``pine family'' to ``a pine tree in the garden'' (Figure \ref{fig:qualitative}).   

Figure \ref{fig:para_incorrect_samples} displays sample outputs with captions that paraphrase the guiding text and sample outputs that use incorrectly labelled guiding text (i.e., the guiding text is not present in the image).

\begin{table}
\tablesize
\begin{center}
\begin{tabular}{l|ccc|c|cc} \hline
$\mathcal{D}$ & \shortI & \shortC & \shortF & Avg & Div-1 & Div-2 \\ \hline \hline
 \data{CC}{GCP} & 0.71 & 0.76 & 0.96 & 0.81 & 0.77 & 0.94 \\
         \data{VG}{GCP} & 0.55 & 0.63 & 0.92 & 0.70 & 0.77 & 0.94\\
         \data{MX}{GCP} & 0.69 & 0.74 & 0.95 & 0.79 & 0.75 & 0.92 \\ \hline \hline
\data{CC}{FRCNN} & 0.73 & 0.69 & 0.96 & 0.79 & 0.76 & 0.94\\
         \data{VG}{FRCNN} & 0.81 & 0.82 & 0.99 & 0.87 & 0.78 & 0.95\\
         \data{MX}{FRCNN} & 0.77 & 0.75 & 0.98 & 0.83 & 0.78 & 0.95\\ \hline
\end{tabular}
\caption {Human evaluation results with the best models from each dataset. These results were trained with \gs extracted using GCP or FRCNN.}
\label{tab:retrain}
\end{center}
\end{table}

\subsection{Train with an Object or Label Detector}
For the results we report in Section \ref{sec:human-eval}, the training \gs are either extracted from the ground truth captions (CC) or from human annotations (VG). 
If we already have access to the upstream model that provides candidate \gs, we can potentially take advantage of this at training time to create a more streamlined model.
For this experiment, we run the FRCNN object and GCP label detector on our training images to obtain training \gs.
To ensure that the ground truth caption is relevant to the \g constructed this way, we only retain the training tuples in which the guiding text is present in the groundtruth caption, using a text-match filter.
For CC, this filter reduces the training tuples from 3.3M to 1.1M when using the FRCNN object detector, and from 3.3M to 2.4M when using the GCP label detector.
For VG, the filter reduces the training tuples from 3.6M to 2.4M when using the FRCNN object detector, and from 3.6M to 1.2M when using the GCP label detector.
Taking the models with the highest \cider score for each dataset and guiding text, we perform another human evaluation and report the scores\footnote{Each instance is evaluated with replication 1, because we found that our human evaluation results were similar between replication 1 and 3: difference was under 0.01 for the majority cases, with one exception where the difference was 0.03.} in Table \ref{tab:retrain}.
We note that the average scores obtained are all higher than their counterparts in Figure \ref{tab:human-eval}, showing that training with a \g distribution that matches the inference-time \g distribution generally leads to better results, e.g., about +8 points across all three dimensions for GCP \gs under CC.

\section{Conclusion}
In this work, we compare Conceptual Captions and Visual Genome to analyze and determine which type of training set performs better 
Guided Image Captioning on out-of-domain data.
We show that although Visual Genome has human-annotated object-level captions and a higher number of unique tokens, training on the Conceptual Captions dataset with web-collected image-level captions that have a high diversity of writing styles and are image-dependent (unlike Visual Genome which has many image-independent target captions) produces better guided captions on out-of-domain data.
Our results suggest that creating a good guided image captioning training set does not require laborious human annotations and that building these datasets by automatically scraping the internet which has the side effect of being more easily scaled-up can lead to better performance.

\section*{Acknowledgments}
We thank Beer Changpinyo and Zhenhai Zhu for discussion and useful pointers during the development of this paper. We thank Ashish Thapliyal for debugging advice and comments on the initial draft of the paper. We would also like to thank Jing Yu Koh for comments on revisions of the paper.

\bibliography{custom}
\bibliographystyle{acl_natbib}

\clearpage
\appendix

\section{Examples of Human Evaluation Form}
\begin{figure}[!ht]
\centering
\includegraphics[width=1.15\linewidth]{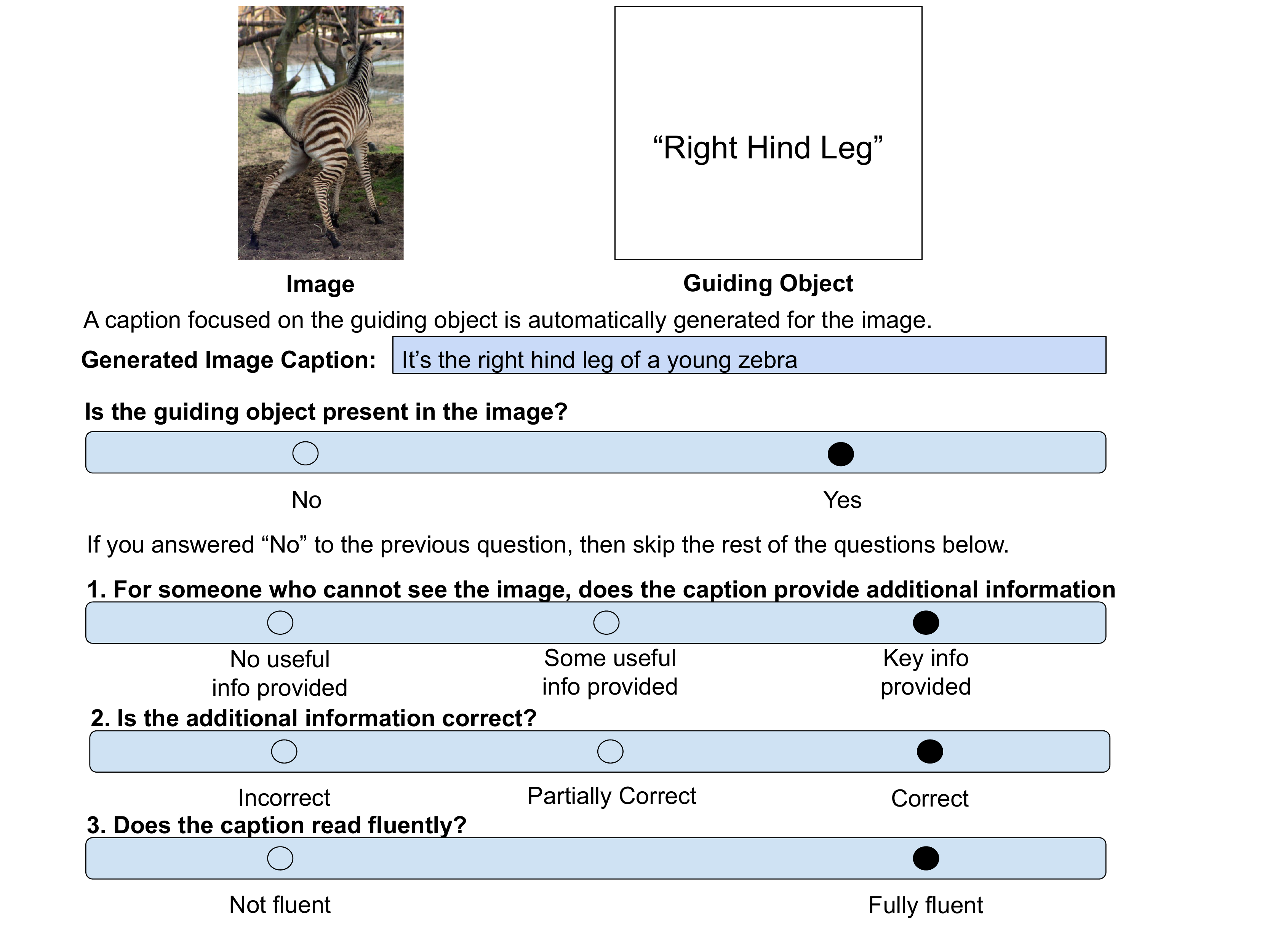}
\caption{Human evaluation form for raters to rate predicted captions.}
\label{fig:human-eval-form-key}
\end{figure}

\begin{figure}[!ht]
\centering
\includegraphics[width=1.15\linewidth]{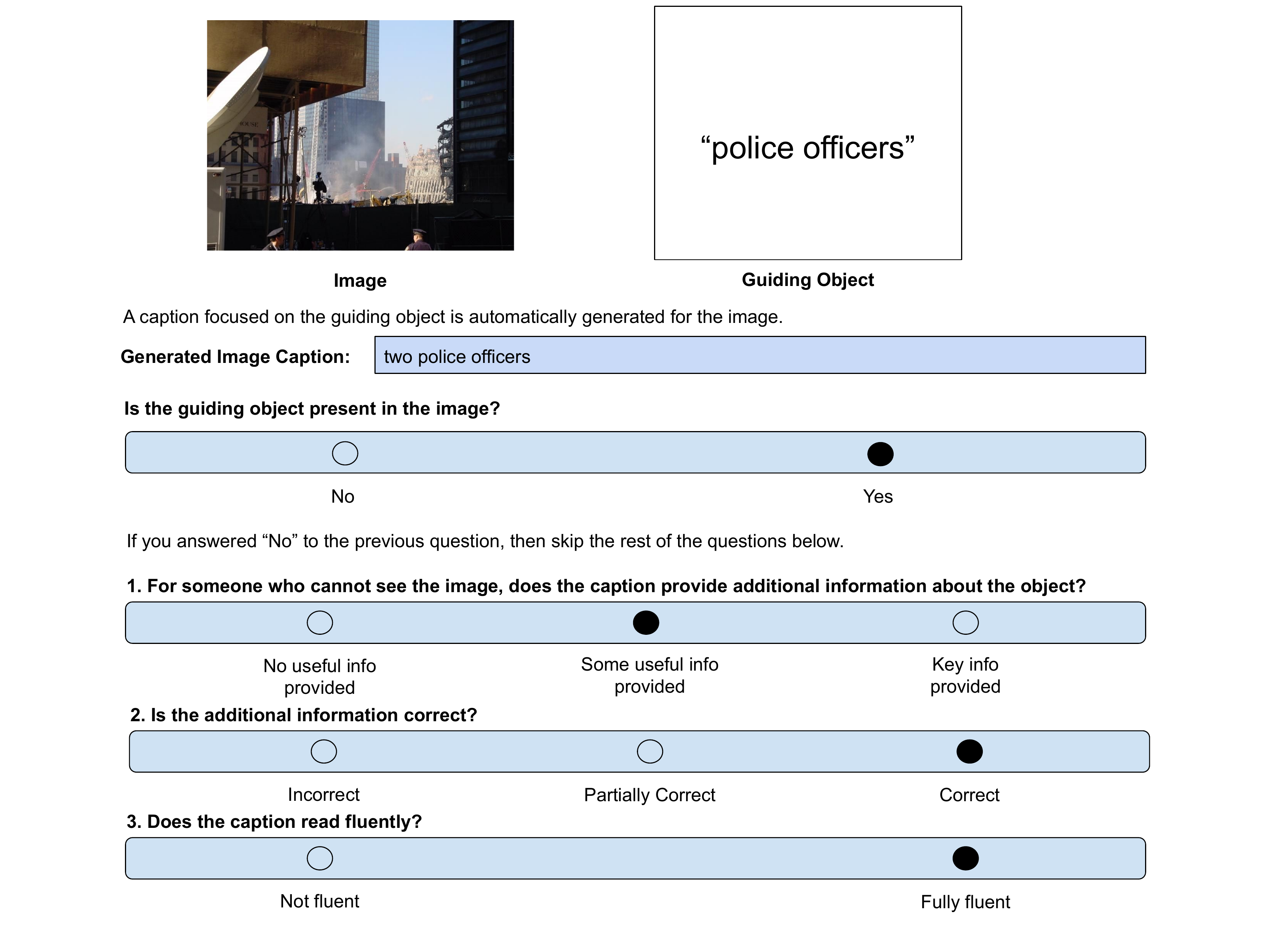}
\caption{Example human evaluation form with caption that should receive partial score for informativeness.}
\label{fig:human-eval-form-partial}
\end{figure}

\end{document}